\documentclass[runningheads]{llncs}
\usepackage{graphicx}
\usepackage{adjustbox}
\usepackage{algorithm, algorithmic}
\usepackage{amsmath, amssymb}
\begin{document}
\title{MSGAT-GRU: A Multi-Scale Graph Attention and Recurrent Model for Spatiotemporal Road Accident Prediction}
\author{Thrinadh Pinjala\inst{1,2} \and
Aswin Ram Kumar Gannina\inst{1,3} \and
Debasis Dwibedy \inst{1,4}}

\institute{Department of Computer Science and Engineering, School of Computing, Coimbatore, Amrita Vishwa Vidyapeetham, India \\ \and
\email{cb.en.u4cse22141@cb.students.amrita.edu}\\ \and
\email{cb.en.u4cse22305@cb.students.amrita.edu}\\ \and 
\email{debasis.dwibedy@gmail.com}\\ 
}
\maketitle              
\begin{abstract}
Accurate prediction of road accidents remains challenging due to intertwined spatial, temporal, and contextual factors in urban traffic. We propose \textsc{MSGAT-GRU}, a multi-scale graph attention and recurrent model that jointly captures localized and long-range spatial dependencies while modeling sequential dynamics. Heterogeneous inputs, such as traffic flow, road attributes, weather, points of interest, are systematically fused to enhance robustness and interpretability. On the Hybrid Beijing Accidents dataset, \textsc{MSGAT-GRU} achieves RMSE = 0.334 and F1-score = 0.878, consistently outperforming strong baselines. Cross-dataset evaluation on METR-LA under a 1-hour horizon further supports transferability, with RMSE = 6.48 (vs. 7.21 for GMAN model) and comparable MAPE. Ablations indicate that three-hop spatial aggregation and a two-layer GRU offer the best accuracy–stability trade-off. These results position \textsc{MSGAT-GRU} as a scalable and generalizable model for intelligent transportation systems, providing interpretable signals that can inform proactive traffic management and road safety analytics.
\end{abstract}
\keywords {Accident Prediction, Gated Recurrent Unit (GRU), Graph Neural Networks (GNN), Road Safety Enhancement, Spatio-temporal Data, Traffic Analysis}
\section{Introduction}\label{sec: Introduction}
Road traffic is an integral part of modern life, supporting economies and connecting communities across the globe. However, road accidents remain a leading cause of injury and death, with the World Health Organization estimating that approximately 1.19 million fatalities occur each year due to road accidents, making it one of the leading causes of mortality worldwide \cite{WHO2023}. As urbanization accelerates and traffic density increases, the need for advanced predictive models capable of anticipating accident-prone scenarios has become more urgent. Effective accident prediction not only helps prevent fatalities but also supports better traffic management, insurance planning, and infrastructure development, thereby contributing to safer and more efficient transportation systems \cite{Partheeban2008}, \cite{Khan2023}.\\\\
\textit{Research Motivation:} Predicting accidents is an inherently complex task due to the interplay of spatial road networks, temporal traffic patterns, weather conditions, and socio-economic factors. Traditional machine learning and neural network models often focus on either spatial relationships or temporal dynamics, limiting their ability to fully capture real-world complexities \cite{Miglani2019}, \cite{AbbasThesis2023}. Furthermore, existing systems frequently struggle with heterogeneous data integration, class imbalance due to the rarity of accidents, and difficulties in adapting to rapidly changing traffic environments \cite{Wu2023}, \cite{Berhanu2023}. These limitations reduce the effectiveness of prediction models when applied to diverse urban contexts or deployed in real-time traffic management systems \cite{Azadani2021}, \cite{Wu2023}.
Motivated by these challenges, our research explores how spatiotemporal graph modeling can enhance accident prediction by integrating diverse data streams and learning patterns across both space and time. By leveraging real-time traffic flow data, meteorological information, road attributes, and calendar-based contextual factors, we aim to address the critical gaps in data heterogeneity, class imbalance, and transferability across cities with different traffic dynamics \cite{Marcillo2022}, \cite{Wang2021}. Our work is inspired by the need for models that not only predict accident risks with high accuracy but also offer interpretable and actionable insights for transportation planners and safety practitioners.\\\\
\textit{Our Contribution:} We propose MSGAT-GRU, an improved model that integrates multi-scale graph attention mechanisms with gated recurrent units (GRU) to effectively capture complex spatial and temporal dependencies in road accident prediction. We evaluate our approach on the Hybrid Beijing Accidents dataset, which includes approximately 7.7 million records, demonstrating significant improvements over existing methods in key metrics such as error reduction and classification performance. Through extensive experiments, including ablation studies and feature importance analysis, we highlight how the integration of heterogeneous data sources and multi-scale modeling enhances both the robustness and interpretability of accident predictions. The results of this study underscore the potential of the proposed model to contribute meaningfully to intelligent transportation systems and road safety interventions. \\\\
\textit{Organization:} The structure of this article is outlined below. Section 2 provides an overview of prior research in road accident prediction, with particular focus on graph-based frameworks and modeling strategies. Section 3 introduces the problem formulation and describes the dataset employed in this work. Section 4 details the architecture and methodological design of the proposed MSGAT-GRU model. Section 5 reports the experimental evaluation, including comparisons with baseline models and an ablation analysis highlighting the role of different components. Finally, Section 6 summarizes the main contributions and discusses potential avenues for future research.
\section{Foundation and Related Work} \label{sec: Foundation and Related Work}
We first review the fundamental concepts underlying spatio-temporal graph modeling, which form the basis for advanced traffic analysis. We then present key contributions from existing graph-based approaches for road traffic and accident prediction, highlighting how they address spatial and temporal dependencies. Finally, we discuss the critical research gaps that persist in the existing literature, emphasizing the need for  more robust, adaptive, and interpretable models.
\subsection{Fundamental Concepts and Definitions} \label{subsec:Fundamental Concepts and Definitions}
\begin{itemize}
\item \textit{Graph Neural Network (GNN):} A class of neural networks designed to operate on graph data structures, consisting of nodes and edges, allowing for information propagation and learning on non-Euclidean data \cite{Scarselli2009}. 
\item \textit{Spatial-Temporal Graph Neural Network (STGNN):} An extension of GNNs that captures both spatial relationships (through graph structure) and temporal dynamics (through sequence modeling) in a unified framework, especially suitable for traffic networks where both dimensions are critical \cite{Yu2018}. 
\item \textit{Graph Attention Network (GAT):} A GNN variant that utilizes attention mechanisms to learn dynamically to assign importance weights to the neighboring nodes while passing messages, allowing for more flexible and context-aware information aggregation \cite{Veličković2018}. 
\item \textit{Gated Recurrent Unit (GRU):} A recurrent neural network (RNN) architecture used for modeling sequential data, employing gating mechanisms (reset and update gates) in order to regulate the flow of information and effectively capture temporal dependencies \cite{Cho2014}. 
\item \textit{Attention Mechanism:} A technique that allows neural networks to focus on specific parts of input data that are most relevant to the task, assigning different weights to different input elements \cite{Bahdanau2015}.
\end{itemize}
\subsection{Related Work in Graph-Based Traffic and Accident Prediction} \label{subsec:Related Work in Graph-Based Traffic and Accident Prediction}
Graph-based frameworks have emerged as effective tools for modeling the intertwined spatial, temporal, and contextual factors in traffic systems. Prior studies can be grouped into five methodological directions:\\
\textit{Dynamic Graph Representations:} Early approaches such as Dynamic-GRCNN \cite{Peng2020} and DSTAGNN \cite{Lan2022} introduced adaptive graph structures to capture evolving traffic flows. While these frameworks effectively model fluctuations, they neglect external contextual variables, such as weather and incidents, limiting their applicability to accident prediction.\\
\textit{Attention-Based Mechanisms:} Models such as TFM-GCAM \cite{Chen2024} and STGNN \cite{Wang2020} employ attention layers to enhance spatiotemporal feature extraction. Although these methods improve short-term forecasting accuracy, their fixed architectures restrict adaptability to rare and unpredictable accident events.\\
\textit{Multi-Graph Approaches:} Frameworks like DMGNN \cite{Ye2023} integrate multiple graphs representing traffic patterns, congestion levels, and accident correlations. Despite improved predictive power, their reliance on handcrafted relationships reduces generalizability across cities with heterogeneous road structures.\\
\textit{Heterogeneous Data Integration:} DSTGCN \cite{Yu2021} and large-scale accident studies by Nippani et al. \cite{Nippani2023} highlight the importance of integrating traffic, meteorological, and POI features. However, their coarse temporal resolution and limited strategies for handling class imbalance undermine real-time accident forecasting.\\
\textit{Network-Wide and Multi-View Learning:} Recent methods such as MSGNN \cite{Tran2023}, STZITD-GNN \cite{Gao2024}, and MG-TAR \cite{Trirat2023} extend prediction to network-wide scales and introduce uncertainty quantification. Yet, these models still lack interpretability and fail to consistently transfer across diverse urban settings.\\
In summary, while graph-based learning has demonstrated clear promise in traffic forecasting, existing approaches are constrained by limited contextual integration, insufficient temporal granularity, and challenges in handling imbalanced accident datasets. These shortcomings motivate the need for more adaptive and interpretable frameworks.
\subsection{Research Gaps}
Building on the above limitations, we identify three critical research gaps that motivate our study in this paper:\\
\textit{(i) Lack of real-time contextual integration:} Most existing models cannot incorporate external factors, such as weather, incidents, and points of interest, in real time. Fixed architectures fail to adapt to the evolving spatiotemporal patterns observed in real-world traffic systems.\\
\textit{(ii) Insufficient temporal resolution:} Many studies aggregate data at coarse daily or monthly levels, overlooking the rapid fluctuations in traffic dynamics. This limits the ability to provide timely, actionable accident predictions.\\
\textit{(iii) Class imbalance in accident datasets:} Since crashes are rare compared to normal traffic flow, severe imbalance skews predictions toward non-accident cases. Current evaluation strategies inadequately address this issue, leading to biased or unreliable outputs.\\
These gaps underscore the need for a robust and flexible framework capable of fusing heterogeneous, fine-grained data, modeling multi-scale spatial dependencies, and mitigating imbalance. Addressing these challenges forms the basis of our proposed MSGAT-GRU model.
\section{Problem Statement}
The objective of this study is to predict the probability of a traffic accident occurring on a road segment $v_i$ during the next time interval $T+1$, given observations up to the current time $T$. We model the road network as an undirected graph 
$G = (V,E,A)$, where $V$ denotes the set of nodes (road segments), $E$ the set of edges (intersections or spatial connections), and $A$ the adjacency matrix encoding the network topology. \\ 
Each node $v_i$ is associated with a feature vector
\[
x_i = \big[x_i^{spatial}, \; x_i^{temporal}, \; x_i^{external}\big],
\]
where $x_i^{spatial}$ includes structural attributes such as road length, number of lanes, and nearby points of interest (POIs),  
$x_i^{temporal}$ captures recent traffic dynamics such as speed and flow patterns, and  
$x_i^{external}$ incorporates contextual information such as weather conditions and calendar-based factors.  \\
The prediction task can be formulated as learning a mapping function:
\[
\hat{y}_i = f\big(x_i^{spatial}, \; x_i^{temporal}, \; x_i^{external}, \; X_{\text{neighbors}}\big),
\]
where $\hat{y}_i \in [0,1]$ represents the probability of an accident on segment $v_i$ at time $T+1$, and $X_{\text{neighbors}}$ denotes features from the $k$-hop neighborhood of $v_i$. \\ 
This formulation must capture four critical aspects:  
(1) spatial dependencies arising from road connectivity and neighboring segments,  
(2) temporal evolution of traffic patterns,  
(3) influence of real-time external factors such as weather and calendar events, and  
(4) the inherent class imbalance in accident datasets.  \\
Addressing these challenges requires a model that integrates multi-scale spatial reasoning, sequential temporal modeling, and heterogeneous feature fusion. The following section presents our proposed methodology and illustrates the MSGAT-GRU model, designed to meet these requirements.
\section{Our Proposed Methodology } \label{sec: Our Proposed Methodology }
This section presents the dataset used in the study, the overall methodology designed to address the problem, and the architecture of the MSGAT-GRU model developed to capture spatial, temporal, and contextual dependencies in accident prediction. We describe how the data sources are integrated and preprocessed to support model training, and how the proposed framework effectively learns from heterogeneous and imbalanced data to improve prediction accuracy.
\subsection{Dataset Description} \label{subsec: Dataset Description}
This study utilizes the Hybrid Beijing Accidents dataset, a comprehensive multi-source traffic dataset designed for spatiotemporal prediction tasks~\cite{Yu2021}. The dataset is publicly available on GitHub\footnote{\url{https://github.com/yule-BUAA/DSTGCN}} and integrates four key components:
\begin{itemize}
\item \textit{Accident records:} Detailed logs of accident events, including location, timestamp, severity, and type of incident.  
\item \textit{Road network data:} Graph representation of the urban road infrastructure, where nodes correspond to road segments and edges capture connectivity through intersections. Attributes such as road type, speed limit, and number of lanes are included.  
\item \textit{Traffic flow data:} Hourly measurements of vehicle count, average speed, and occupancy, providing both real-time and historical traffic dynamics.  
\item \textit{External contextual data:} Weather conditions (temperature, precipitation, visibility, wind speed, and weather events) and point-of-interest (POI) information (schools, shopping centers, hospitals, etc.), capturing environmental and socio-economic influences on traffic safety.  
\end{itemize}
The dataset spans a three-month period (August–November 2018) and includes millions of traffic and contextual records. To mitigate the inherent class imbalance, we adopt a balanced sampling strategy during training, maintaining an equal ratio of accident to non-accident instances. Extensive preprocessing aligns heterogeneous sources across space and time, ensuring that each road segment is represented as a node enriched with static attributes and dynamic features.\\
This dataset is well-suited for our study as it provides heterogeneous, fine-grained information necessary to capture the spatial, temporal, and contextual dependencies that govern accident risk in large-scale urban environments.
\subsection{Various Phases of Our Methodology}
\begin{figure}[h]
    \centering
    \includegraphics[width=1.0\textwidth]{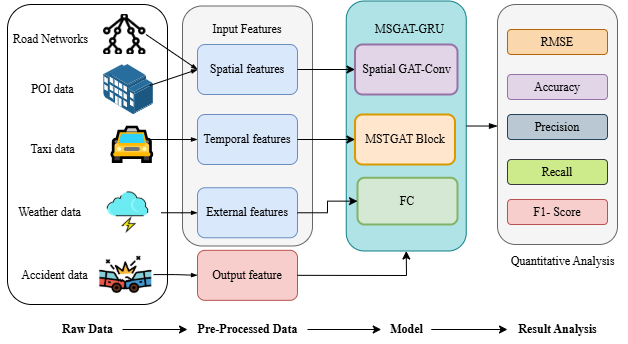}
    \caption{Workflow of the proposed methodology, highlighting data integration, preprocessing, balanced sampling, and dataset partitioning prior to model training.}
    \label{fig:methodology}
\end{figure}

The proposed methodology follows a systematic pipeline, illustrated in Figure \ref{fig:methodology}, which integrates heterogeneous data, ensures consistency through preprocessing, and prepares structured inputs for the MSGAT-GRU model.
\\
\textit{Data Integration:} The first phase constructs meaningful feature representations for each road segment by integrating three modalities: (i) \emph{spatial features} combining graph-based road network attributes with point-of-interest (POI) density, (ii) \emph{temporal features} derived from traffic speed sequences resampled at hourly intervals, and (iii) \emph{external features} incorporating weather conditions and calendar-based indicators. This multimodal fusion provides a rich contextual basis for accident prediction.\\
\textit{Preprocessing:} To ensure consistency across heterogeneous sources, several preprocessing steps are applied. Road segment attributes are standardized to maintain geometric and topological coherence. Traffic speed data, originally irregular, is aligned through optimized interpolation within $\pm7$ or $\pm14$ day windows, with mean imputation as fallback. All feature groups are normalized independently to maintain scale balance and improve convergence stability. For each accident event, a $k$-hop subgraph is dynamically extracted, enabling localized spatiotemporal context representation.\\
\textit{Balanced Sampling:} Accident datasets are inherently imbalanced. To mitigate bias, positive samples are derived from real accident records while negative samples are selected from non-accident instances under comparable spatial constraints. Both are validated for geographic consistency to ensure representative learning.\\
\textit{Dataset Partitioning.} Finally, the dataset is split into training (70\%), validation (10\%), and test (20\%) subsets. This stratification minimizes selection bias and supports robust evaluation of model generalization.\\
Overall, this pipeline provides a structured and balanced representation of heterogeneous traffic environments, forming the foundation for the subsequent MSGAT-GRU model design.
\subsection{Our Proposed MSGAT-GRU Model}
\begin{figure}[h]
    \centering
    \includegraphics[width=1.0\textwidth]{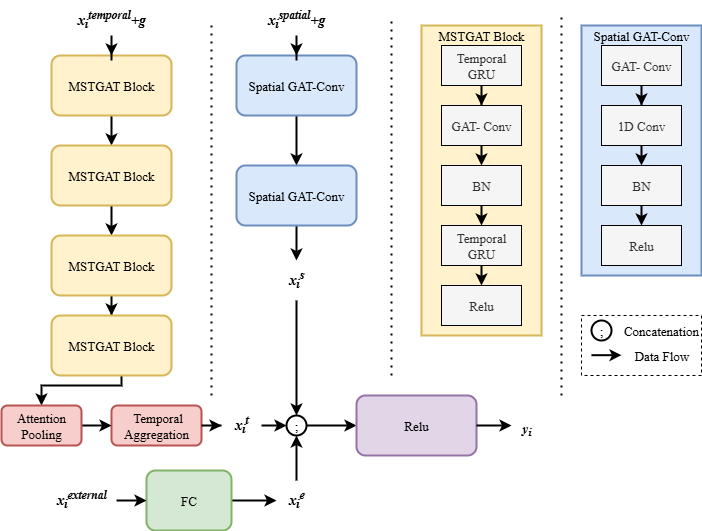}
    \caption{Architecture of the proposed MSGAT-GRU model. Spatial, temporal, and external features are processed through dedicated branches with graph attention and recurrent units before feature fusion for accident risk prediction.}
    \label{fig:model}
\end{figure}

The MSGAT-GRU model is designed to capture complex spatiotemporal dependencies by processing spatial, temporal, and external features through specialized branches before fusing them into a unified representation. Figure~\ref{fig:model} illustrates the overall architecture.\\
\textit{Input Features and Branch Separation:} Each road segment is characterized by three feature sets: (i) \emph{spatial features}, encoding topological road attributes and points of interest, (ii) \emph{temporal features}, representing historical traffic dynamics, and (iii) \emph{external features}, such as weather and calendar data. These are processed independently to preserve modality-specific characteristics.\\\\
\textit{Spatial Feature Processing:} Spatial features are passed through stacked Graph Attention (GAT) layers, enabling node-level embeddings that account for heterogeneous neighbor influence. For node $i$, the representation is updated as:
\[
h'_i = \sigma \left(\sum_{j \in \mathcal{N}(i)} \alpha_{ij} W h_j \right),
\]
where $\alpha_{ij}$ is the attention coefficient learned via normalized importance scores. This dynamic weighting allows the model to prioritize influential neighbors over uniform aggregation.\\\\
\textit{Temporal Feature Processing:} Temporal sequences are modeled using Multi-Scale Spatiotemporal Graph Attention (MSTGAT) blocks. Each block combines bi-directional GRUs, which capture sequential dependencies, with multi-scale GAT modules aggregating information from 1-hop, 2-hop, and 3-hop neighbors:\\
\[
h^{\text{multi-scale}}_i = \big\|_{s=1}^S h^{(s)}_i,
\]
where $\|$ denotes concatenation across scales. This design preserves both local and global traffic dependencies without over-smoothing.\\\\
\textit{External Feature Processing:} External factors are embedded through a lightweight fully connected (FC) network with normalization and ReLU activation:
\[
z_{\text{external}} = \text{FC}(\text{BN}(\text{ReLU}(\text{Linear}_1(x_{\text{external}})))).
\]
This ensures compact yet informative representations of weather and calendar influences.\\\\
\textit{Attention Pooling and Temporal Aggregation:} To derive graph-level embeddings, attention pooling assigns learnable weights $\beta_i$ to nodes, emphasizing their relative importance:
\[
h_{\text{temp-agg}} = \sum_{i=1}^{N} \beta_i h_i.
\]
The resulting sequence is passed through a bidirectional GRU, enabling the model to jointly capture short- and long-term temporal patterns.\\\\
\textit{Feature Fusion and Prediction:} Finally, outputs from the three branches are concatenated:
\[
h_{\text{final}} = [h_{\text{temp}} \,\|\, h_{\text{spatial}} \,\|\, z_{\text{external}}],
\]
and passed through an FC layer with dropout and normalization to predict accident probability $\hat{y}$. This fusion strategy ensures comprehensive integration of spatial, temporal, and contextual dependencies.\\\\
The modular design of MSGAT-GRU provides three advantages: (i) fine-grained modeling of heterogeneous data sources, (ii) preservation of multi-scale dependencies across space and time, and (iii) interpretable representations that highlight influential road segments and contextual factors. The key notations and mathematical definitions used in this architecture are summarized in Table 1.
\begin{table}[ht]
\centering
\caption{Notations and Definitions used in the MSGAT-GRU Model}
\scriptsize
\begin{adjustbox}{width=\linewidth}
\begin{tabular}{lp{11cm}}
\hline
\textbf{Notation} & \textbf{Definition} \\
\hline
$G_t = (V, E, X_t)$ & raph at time $t$, with nodes $V$, edges $E$, and node feature matrix $X_t$. \\
$V$ & Set of nodes representing road segments in the road network. \\
$E$ & Set of edges representing connections between adjacent road segments. \\
$X_t \in \mathbb{R}^{|V| \times F}$ & Node feature matrix at time $t$, with $F$ features per node. \\
$x_t^i \in \mathbb{R}^F$ & Feature vector for node $v_i$ at time $t$, including spatial, temporal, and external features. \\
$p_{t+\Delta t}^i$ & Predicted probability of a traffic accident at node $v_i$ within future time window $\Delta t$. \\
$T$ & Number of historical time steps used for prediction. \\
$f$ & Function mapping historical graph sequence to accident probabilities. \\
$x_i^{\text{spatial}}$ & Spatial features for node $v_i$, combining graph network data and POI data. \\
$x_i^{\text{temporal}}$ & Temporal features for node $v_i$, derived from traffic speed data. \\
$x_i^{\text{external}}$ & External features for node $v_i$, including weather and calendar information. \\
$x_i^p$ & Point of Interest (POI) data for node $v_i$. \\
$x_i^g$ & Graph network data for node $v_i$ (e.g., road type, number of lanes). \\
$x_i^w$ & Weather attributes for node $v_i$ (e.g., temperature, precipitation). \\
$x_i^c$ & Calendar features for node $v_i$ (e.g., day of week, hour). \\
$y$ & Binary label indicating accident occurrence at a specific time and location. \\
$A$ & Adjacency matrix defining the road network topology. \\
$h'_i$ & Updated feature representation for node $v_i$ after GAT convolution. \\
$\alpha_{ij}$ & Normalized attention score between nodes $v_i$ and $v_j$ in GAT. \\
$h_t$ & Hidden state of the GRU at time $t$ for temporal feature processing. \\
$h^{\text{multi-scale}}_i$ & Concatenated multi-scale spatial representation for node $v_i$. \\
$z^{\text{external}}$ & Processed external feature embedding after fully connected layers. \\
$\beta_i$ & Attention score for node $v_i$ in temporal feature pooling. \\
$h^{\text{temp-agg}}$ & Aggregated graph-level temporal representation. \\
$h^{\text{spatial}}$ & Pooled spatial feature representation across nodes. \\
$h^{\text{final}}$ & Final concatenated feature vector for accident probability prediction. \\
$\hat{y}$ & Predicted accident probability output by the model. \\
\hline
\end{tabular}
\end{adjustbox}
\end{table}
\section{Performance Evaluation} \label{sec: Performance Evaluation}
This section presents the experimental evaluation of the proposed MSGAT-GRU model on the Hybrid Beijing dataset using widely accepted regression and classification metrics, including RMSE, Accuracy, Precision, Recall, and F1-score. The obtained results are compared against established baseline models to demonstrate the effectiveness of our approach. Additionally, we assess the model’s performance on the METR-LA dataset by employing standard error metrics such as MAE, RMSE, and MAPE, and benchmark it against the state-of-the-art GMAN model. Furthermore, an ablation study is conducted to investigate the influence of key architectural parameters, specifically, the GAT hop size and GRU depth, on prediction performance. Finally, the computational cost associated with the model’s design is analyzed to provide insights into its practical applicability.
\subsection{Result Comparison and Discussions} \label{subsec: Result Comparison and Discussions}
We conducted extensive experiments on the Hybrid Beijing Accidents dataset using a 60-minute prediction horizon across diverse traffic settings, including urban, suburban, and rural road segments. The MSGAT-GRU model was benchmarked against prominent baselines, including DSTGCN, MSGCN, STGAT-GRU, and STGCN variants. Table \ref{tab:performance_comparison} summarizes the comparative performance.
\begin{table}[ht]
\centering
\caption{Performance comparison of MSGAT-GRU with baseline models on the Beijing dataset for 60-minute accident prediction}
\scriptsize
\begin{adjustbox}{width=\linewidth}
\begin{tabular}{lccccc}
\hline
\textbf{Model} & \textbf{RMSE} & \textbf{Accuracy} & \textbf{Precision} & \textbf{Recall} & \textbf{F1-Score} \\
\hline
STGAT & $0.378 \pm 0.015$ & $0.814 \pm 0.012$ & $0.831 \pm 0.013$ & $0.762 \pm 0.014$ & $0.795 \pm 0.010$ \\
STGCN\_GraphSAGE & $0.370 \pm 0.011$ & $0.843 \pm 0.009$ & $0.777 \pm 0.012$ & $0.785 \pm 0.013$ & $0.781 \pm 0.008$ \\
STGCN\_LSTM & $0.361 \pm 0.013$ & $0.801 \pm 0.011$ & $0.789 \pm 0.010$ & $0.798 \pm 0.012$ & $0.794 \pm 0.009$ \\
STGAT\_GRU  & $0.344 \pm 0.012$ & $0.812 \pm 0.008$ & $0.824 \pm 0.011$ & $0.893 \pm 0.009$ & $0.857 \pm 0.007$ \\
MSGCN            & $0.357 \pm 0.009$ & $0.878 \pm 0.006$ & $0.835 \pm 0.008$ & $0.817 \pm 0.010$ & $0.826 \pm 0.005$ \\
MSGCN\_GRU  & $0.352 \pm 0.008$ & $0.883 \pm 0.005$ & $0.841 \pm 0.007$ & $0.828 \pm 0.009$ & $0.834 \pm 0.006$ \\
DSTGCN      & $0.343 \pm 0.011$ & $0.867 \pm 0.007$ & $0.821 \pm 0.0012$ & $\mathbf{0.896 \pm 0.016}$ & $0.8508 \pm 0.011$ \\
\hline
\textbf{MSGAT-GRU (Ours)} & $\mathbf{0.334 \pm 0.016}$ & $\mathbf{0.886 \pm 0.004}$ & $\mathbf{0.896 \pm 0.005}$ & $0.861 \pm 0.006$ & $\mathbf{0.878 \pm 0.004}$ \\
\hline
\end{tabular}
\label{tab:performance_comparison}
\end{adjustbox}
\end{table}\\
MSGAT-GRU consistently outperforms baselines across both regression and classification metrics. It reduces RMSE to 0.334, improving upon DSTGCN by approximately 1\%, and achieves an F1-score of 0.878, a 2.1\% gain over the strongest competitor. While these numerical margins may appear modest, they are significant in highly imbalanced accident datasets where even small improvements translate into better identification of rare events and fewer false alarms. Figure 3 illustrates performance across prediction horizons. As expected, accuracy decreases with longer horizons due to higher uncertainty. Nevertheless, MSGAT-GRU maintains relatively stable performance even at 60 minutes, outperforming all baselines in both RMSE and F1-score.\\\\
\begin{figure}[h]
\centering
\includegraphics[width=1.1\textwidth]{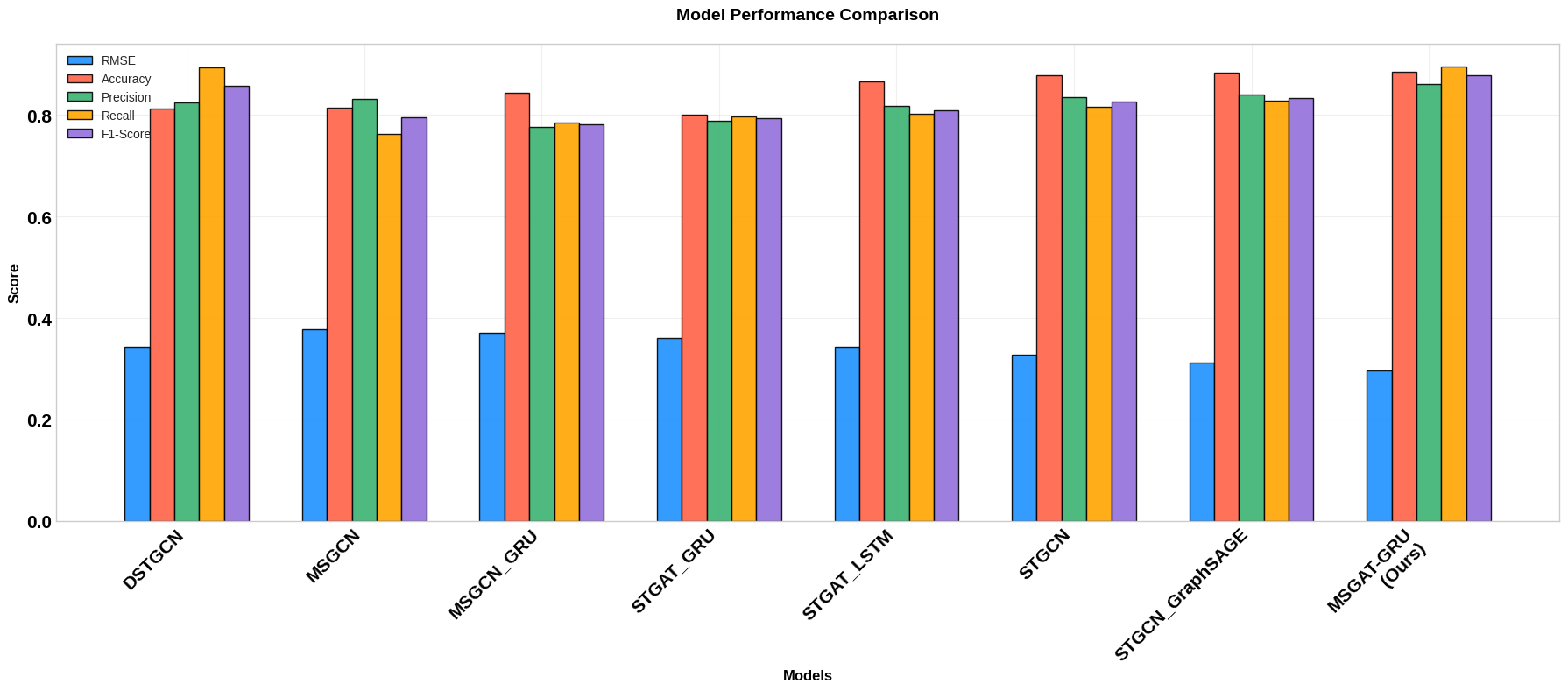}
\caption{Comparison of RMSE, Accuracy, Precision, Recall, and F1-score across different prediction horizons. MSGAT-GRU consistently outperforms baseline models, particularly at longer horizons where uncertainty is higher.}
\end{figure}
\textit{Discussion:} The improved performance stems from three design choices. First, the multi-scale GAT captures hierarchical spatial correlations (1-hop to 3-hop neighborhoods) more effectively than single-scale baselines. Second, bi-directional GRUs model sequential dynamics across time, improving recall without sacrificing precision. Third, integrating external features ensures that contextual factors such as weather and calendar events are explicitly considered. Together, these components yield a balanced improvement across regression and classification metrics, underscoring the robustness of the framework for accident prediction.
\subsection{Ablation Study: Impact of GAT Hop and GRU Depth}
\begin{figure}[h]
    \centering
    \includegraphics[width=1.0\textwidth]{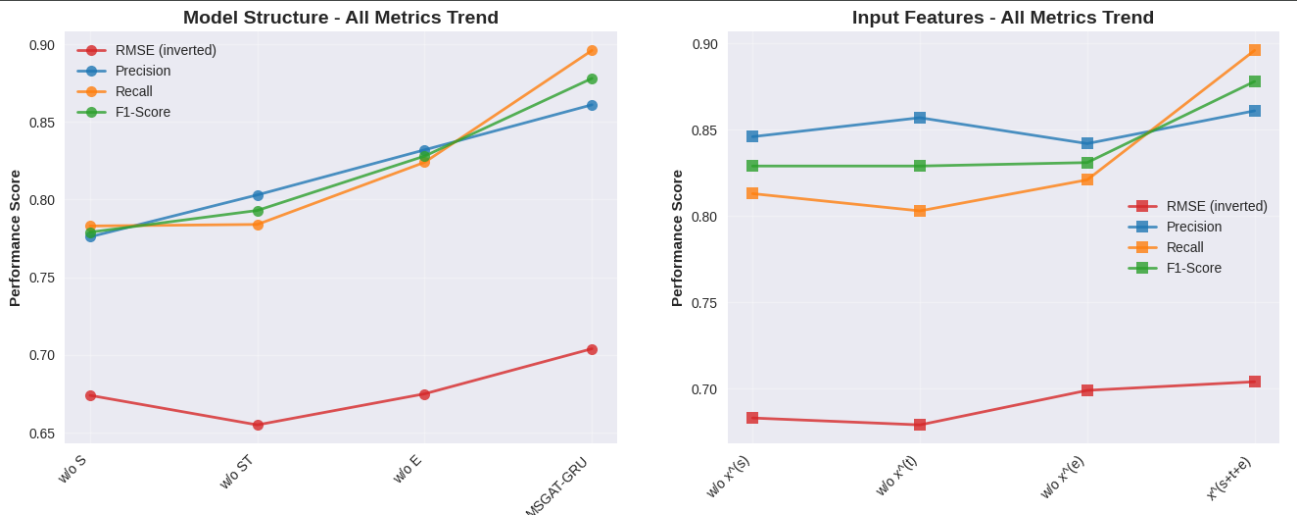}
    \caption{Performance trends under varying GAT hop sizes and GRU depths. Multi-hop aggregation improves spatial learning, while two-layer GRUs achieve the best balance between predictive accuracy and stability.}
    \label{fig:ablation}
\end{figure}

To better understand the contributions of key architectural choices, we conducted an ablation study varying two hyperparameters: (i) the number of Graph Attention (GAT) hops and (ii) the depth of Gated Recurrent Units (GRU). The objective was to evaluate how spatial aggregation scope and temporal modeling depth influence accident prediction performance. The results are summarized in Table \ref{tab:ablation_gat_gru}.
\begin{table}[ht]
    \centering
    \caption{Ablation study results under different GAT hop configurations and GRU depths on the Beijing dataset. Best results in bold.}
    \label{tab:ablation_gat_gru}
\begin{tabular}{cccccc}
\hline
        GAT Hops & GRU Depth & RMSE & Precision & Recall & F1-Score \\
        \hline
        1      & 1 & 0.351 & 0.837 & 0.801 & 0.819 \\
        1-2    & 1 & 0.342 & 0.846 & 0.814 & 0.830 \\
        1-2-3  & 1 & 0.341 & 0.850 & 0.828 & 0.839 \\
        \hline
        1      & 2 & 0.344 & 0.845 & 0.816 & 0.830 \\
        1-2    & 2 & 0.338 & 0.854 & 0.832 & 0.843 \\
        \textbf{1-2-3}  & \textbf{2} & \textbf{0.334 }& \textbf{0.896} & \textbf{0.861} & \textbf{0.878} \\
        \hline
        1      & 3 & 0.349 & 0.842 & 0.807 & 0.824 \\
        1-2    & 3 & 0.347 & 0.848 & 0.823 & 0.835 \\
        1-2-3  & 3 & 0.339 & 0.858 & 0.887 & 0.872 \\

\hline
    \end{tabular}
\end{table}\\
\textit{Effect of GAT Hop Sizes:} Extending the receptive field from single-hop to multi-hop neighborhoods improves spatial representation learning. Incorporating 1–2–3 hops enables the model to capture both local interactions (e.g., immediate neighbors) and broader network-wide correlations (e.g., downstream congestion effects). This hierarchical aggregation consistently reduces RMSE and enhances F1-scores, confirming the benefit of multi-scale spatial reasoning.\\
\textit{Effect of GRU Depth:} Increasing GRU layers improves temporal modeling up to a point. Two-layer GRUs provide the best trade-off between expressiveness and stability, yielding the strongest overall performance. Deeper configurations (three layers) show diminishing returns and occasional instability, likely due to overfitting and increased computational cost.\\
\textit{Key Insight:} The best configuration, three-hop GAT aggregation with two-layer GRUs, achieves an RMSE of 0.334 and F1-score of 0.878. This balance between spatial context and temporal depth is further visualized in Figure \ref{fig:ablation}, which highlights performance trends across different model variants. The figure illustrates that multi-hop aggregation provides robust spatial learning, while two-layer GRUs offer the optimal trade-off between accuracy and stability.
\subsection{Comparison with State-of-the-Art Model on METR-LA Dataset}
\label{subsec:comparison_metrla}
To further assess the generalizability of MSGAT-GRU beyond the Hybrid Beijing dataset, we benchmarked it on the widely used METR-LA traffic forecasting dataset under a 1-hour prediction horizon, following standard spatio-temporal evaluation protocols. The state-of-the-art Graph Multi-Attention Network (GMAN) \cite{Zheng2020} was selected as the comparison baseline, given its strong performance in modeling spatial dependencies through attention mechanisms. 
\begin{table}[ht]
    \centering
    \caption{Performance comparison between MSGAT-GRU and GMAN on the METR-LA dataset for 1-hour prediction}
    \label{tab:metrla_comparison}
    \begin{tabular}{lccc}
        \hline
        Model & MAE & RMSE & MAPE (\%) \\
        \hline
        GMAN & 3.40 & 7.21 & 9.72 \\
        MSGAT-GRU & 3.43 & \textbf{6.48} & \textbf{9.69} \\
        \hline
    \end{tabular}
\end{table}
As shown in Table \ref{tab:metrla_comparison}, MSGAT-GRU attains a lower RMSE (6.48 vs. 7.21) and a marginally better MAPE (9.69\% vs. 9.72\%), while exhibiting a slightly higher MAE (3.43 vs. 3.40). These results suggest that although GMAN better minimizes small deviations, MSGAT-GRU is more effective in controlling large errors, critical for accident prediction, where rare but severe deviations can disproportionately affect safety outcomes. The ability of MSGAT-GRU to aggregate spatial cues across multiple hops and simultaneously encode temporal dynamics allows it to adapt more robustly to abrupt fluctuations and noise inherent in real-world urban traffic.\\
Overall, the transfer learning results on METR-LA reinforce the robustness of MSGAT-GRU across diverse datasets, complementing the improvements observed in Section 5.2. This demonstrates the model’s capability to generalize beyond city-specific conditions and underscores its potential for deployment in heterogeneous traffic environments. 
\subsection{Inferences}
The experimental results collectively highlight how MSGAT-GRU addresses key challenges in spatio-temporal accident prediction. Multi-scale graph attention captures both localized and long-range dependencies, while recurrent modeling preserves sequential patterns, leading to consistent improvements in recall and F1-score on imbalanced accident data. The ablation study further confirmed that a three-hop receptive field combined with a two-layer GRU yields the optimal balance between accuracy and stability. The cross-dataset validation on METR-LA provided additional evidence of generalizability, showing that the model remains robust under traffic conditions markedly different from Beijing. This indicates that the fusion of heterogeneous features, including traffic flow, weather, and points of interest, not only enhances prediction within a single domain but also facilitates transferability across diverse urban settings.\\
Taken together, these findings establish MSGAT-GRU as a scalable and interpretable framework for accident prediction. By effectively unifying spatial reasoning, temporal sequence modeling, and contextual integration, the model sets a strong foundation for advancing intelligent transportation systems and for supporting real-time traffic safety interventions.
\subsection{Computation Cost Analysis}
\label{subsec:cost}
All experiments were conducted on a machine with an Intel Core i7-12700H, NVIDIA GeForce RTX 3050 Ti, and 16GB RAM. With the default training configuration, each epoch took approximately 10 minutes. This overhead primarily stems from the \emph{multi-scale} graph attention used in the architecture (see Figure \ref{fig:model}), which aggregates information across multiple hop distances in parallel rather than relying on single-hop propagation.\\ \footnote{The MSTGAT blocks are depicted in the model architecture (Figure \ref{fig:model}); they are not part of the data workflow in Figure \ref{fig:methodology}.}
To contextualize cost, let $N$ be the number of nodes, $E$ the number of edges, $T$ the look-back length, $S$ the number of spatial scales (hops), $H$ the number of attention heads, and $d$ the hidden size. A single MSGAT layer incurs
\[
\mathcal{O}\!\big(SH\cdot(Ed + Nd^2)\big)
\]
per time step due to edge-wise message passing and node-wise transformations; the temporal GRU contributes 
\[
\mathcal{O}\!\big(T\,N d^2\big).
\]
Hence, inference scales \emph{linearly} with $S$ and $H$, \emph{near-linearly} with $E$ for sparse graphs, and \emph{linearly} with $T$. In practice, the transition from single-hop to three-hop aggregation increases spatial cost roughly proportionally to $S$, which is consistent with the accuracy gains observed in the ablation (Table \ref{tab:ablation_gat_gru}). \\
From a deployment standpoint, three techniques were most beneficial: (i) sparse adjacency storage to reduce memory bandwidth, (ii) moderate hidden sizes to control $Nd^2$ terms, and (iii) caching static embeddings for frequently queried segments. While multi-scale attention introduces overhead relative to single-hop baselines, the \textit{accuracy improvements} reported earlier (Tables \ref{tab:performance_comparison} and \ref{tab:ablation_gat_gru}) provide a favorable accuracy–cost trade-off for safety-critical applications.
\section{Conclusion}
\label{sec:conclusion}
We proposed \textsc{MSGAT-GRU}, a multi-scale graph attention and recurrent model for spatio-temporal accident prediction that fuses traffic flow, road-network attributes, weather, and points of interest. On the Hybrid Beijing Accidents dataset, the model achieves RMSE = 0.334 and F1-score = 0.878, outperforming strong graph-based baselines across both regression and classification metrics. Cross-dataset evaluation on METR-LA (1-hour horizon) further indicates robust transferability with a lower RMSE = 6.48 than GMAN (7.21), while maintaining comparable MAPE. Ablations confirm that three-hop spatial aggregation coupled with a two-layer GRU provides the best accuracy–stability trade-off.\\
\textit{Limitations:} Balanced sampling improves learning on imbalanced data but may shift operating points in deployment where accidents are rare. Multi-scale attention introduces additional computational overhead, which can be constraining under tight latency or edge-resource budgets.\\
\textit{Outlook:} Future work will explore (i) \emph{lightweight} variants (e.g., sparse or neighborhood-sampled attention, parameter sharing, and quantization/pruning), (ii) \emph{explainability and calibration} for risk-aware decision support, and (iii) integration of real-time connected-vehicle and driver-behavior streams for finer-grained, short-horizon risk estimation. These directions aim to retain the demonstrated accuracy while improving deployability and trust in intelligent transportation systems.

\end{document}